# YM-WML: A new Yolo-based segmentation Model with Weighted Multi-class Loss for medical imaging

Haniyeh Nikkhah
Department of electrical and computer engineering
University of Tabriz
Tabriz, Iran
h.nikkhah1401@ms.tabrizu.ac.ir

Jafar Tanha
Department of electrical and computer engineering
University of Tabriz
Tabriz, Iran
tanha@tabrizu.ac.ir

Mahdi Zarrin
Department of electrical and computer engineering
University of Tabriz
Tabriz, Iran
mahdi.zarrin1401@ms.tabrizu.ac.ir

SeyedEhsan Roshan
Department of electrical and computer engineering
University of Tabriz
Tabriz, Iran
ehsan.roshan@tabrizu.ac.ir

Amin Kazempour
Department of electrical and computer engineering
University of Tabriz
Tabriz, Iran
amin.kazempour1402@ms.tabrizu.ac.ir

*Abstract*— Medical image segmentation poses significant challenges due to class imbalance and the complex structure of medical images. To address these challenges, this study proposes YM-WML, a novel model for cardiac image segmentation. The model integrates a robust backbone for effective feature extraction, a YOLOv11 neck for multi-scale feature aggregation, and an attention-based segmentation head for precise and accurate segmentation. To address class imbalance, we introduce the Weighted Multi-class Exponential (WME) loss function. On the ACDC dataset, YM-WML achieves a Dice Similarity Coefficient of 91.02, outperforming state-of-the-art methods. The model demonstrates stable training, accurate segmentation, and strong generalization, setting a new benchmark in cardiac segmentation tasks.

## I. INTRODUCTION

Image segmentation is a key task in computer vision, where each pixel in an image is assigned a specific label. This approach provides a deeper understanding of the image than conventional image processing methods [1]. It plays a crucial role in diverse fields, including image and video analysis, autonomous vehicles, and medical imaging.

In the medical field, image segmentation is crucial for identifying abnormal tissues, such as tumors, lesions, or organs, in medical images like X-rays, CT scans, and MRI. This helps determine the shape, volume, and location of regions of interest, aiding in disease diagnosis and treatment planning [2]. Accurate segmentation is essential for assessing the size and spread of lesions, which supports treatment planning and monitoring. Semantic segmentation becomes even more critical in multi-class medical imaging tasks, where multiple structures need to be segmented simultaneously [4]. These tasks present challenges such as overlapping structures and varying shapes, requiring robust segmentation algorithms [3].

In cardiac imaging, semantic segmentation plays a key role in identifying structures like the left ventricle, right ventricle, and myocardium in MRI scans. These tasks are complicated by variations in size, shape, and tissue characteristics, particularly in conditions such as myocardial infarction or cardiomyopathy. Effective segmentation is critical for accurate diagnosis and treatment planning, especially in complex cases with unclear boundaries and varying tissue thickness. By addressing these challenges, semantic segmentation enhances diagnosis, treatment planning, and disease monitoring across various medical imaging tasks [3].

Recent advancements in deep learning have significantly improved medical image segmentation [5–8]. Convolutional neural networks (CNNs) are widely used due to their ability to learn complex patterns from medical images [9], effectively segmenting structures such as organs, tumors, and blood vessels. Common models like U-Net [10], SegNet [11], and FCN [9] are effective in segmenting medical images. U-Net, for instance, uses an encoder-decoder architecture with skip connections to classify pixels for accurate segmentation. In addition to traditional models, YOLO (You Only Look Once) [12] has shown promise in medical image analysis. Known for its real-time object detection, YOLO is effective in identifying and localizing medical anomalies, making it a powerful tool for segmentation. Its ability to quickly detect and localize structures enhances segmentation accuracy and speed, making it valuable for medical diagnostics [13]. When used for segmentation, YOLO's capabilities in detecting and localizing features allow for efficient and precise analysis of medical images, improving both speed and accuracy in medical diagnostics [13,14].

Selecting the right loss function is critical for optimizing semantic segmentation in deep learning models, especially in medical images, where class imbalance is a key challenge [15]. Various loss functions have been proposed, including distribution-based (e.g., focal and cross-entropy loss), region-based (e.g., dice loss), boundary-based (e.g., boundary loss)

[16], and compound losses (e.g., combination of dice and cross-entropy loss) [17]. While cross-entropy, dice loss, and focal loss are commonly used in medical segmentation tasks, most of these do not specifically address class imbalance, which is crucial in medical image segmentation [15].

The proposed contributions of this study are summarized as follows:

- A novel loss function, called WME, is introduced to address the class imbalance issue in medical imaging segmentation.
- A new model based on YOLOv11 with a redesigned backbone is developed.
- Comprehensive experiments are conducted to demonstrate the superiority of the proposed model and loss function.

The structure of the study is organized as follows: first, a review of related works and previous methods is presented. Second, the proposed approaches are detailed. Finally, the experimental results showcasing the performance of the proposed method are discussed.

## II. RELATED WORKS

The most commonly adopted model architectures for medical image segmentation are Convolutional Neural Networks (CNNs), due to the ability of CNNs to capture spatial relationships in images more efficiently than traditional pattern recognition algorithms [19]. Nonetheless, they face difficulties in modeling long-range dependencies among pixels, which has resulted in recent inclusion of attention mechanisms and transformer-based models. Early hybridization explored the effective utilization of CNNs and attention, including Attention UNet [20] and TransUNet [21], which integrated attention mechanisms for better propagation of high-performance CNN features and for modeling long-range context. The integration has been further achieved with modern architectures such as [22] and nnUNet [23], allowing for a better contextual understanding and automation of the fitting process to the segmentation task. The self-attention mechanism of Transformers has shown great promise in the medical image segmentation domain as they overcome the limitations of CNNs by capturing long-range dependencies in sequential data. Vision Transformers (ViTs) [24] and some of their better-updated derivatives such as the Swin Transformer [25] and MaxViT [26] can capture global and local features to a larger extent, such techniques show potential for providing better accuracy on segmentation. Hybrid methods that leverage the benefits of both CNNs and transformers have emerged as state-of-the-art segmentation tools, particularly for challenging medical imaging and pathology applications.

## III. METHODOLOGY

This methodology is divided into two parts: the architecture, which includes the backbone, YOLOv11 neck, and attention-based segmentation head (SAH), and the proposed weighted multi-class exponential (WME) loss function, designed to address class imbalance.

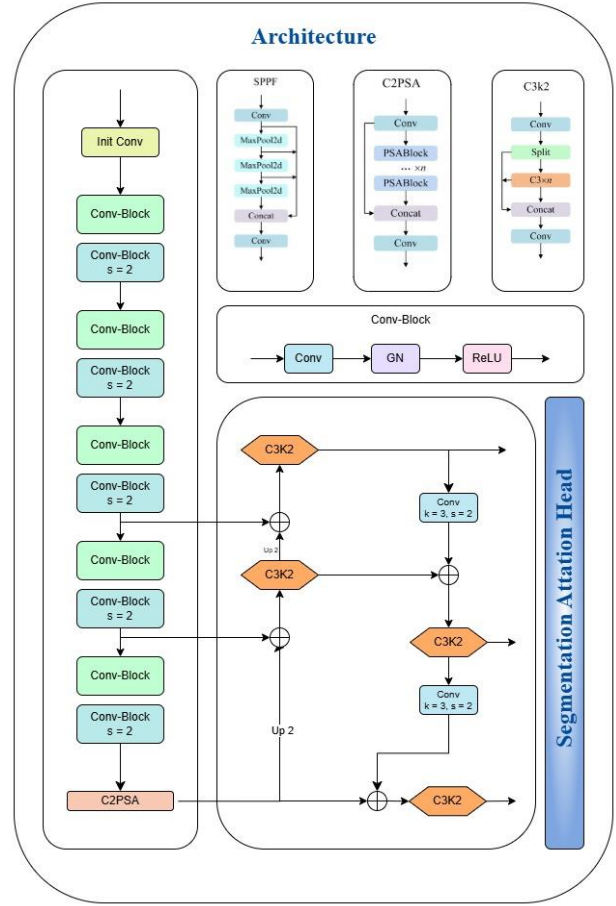

Fig. 1. YM-WML Architecture

*1. Architecture*

In this paper, we propose an architecture for the semantic segmentation of medical images with two or more classes. The feature extraction component (backbone) is designed using convolutional layers tailored to the specific requirements of the task. To enhance and aggregate the feature maps from the backbone, we incorporate the YOLOv11 neck as an intermediate module. The output of the neck is then passed to the segmentation head, which is uniquely designed with attention mechanisms to perform the segmentation task effectively. In the following sections, we will explain each component of the architecture in detail, including the backbone, the neck, and the attention-based segmentation head.

## A. Backbone

In our proposed architecture, the Backbone is responsible for extracting meaningful features from the input medical images. As shown in Figure 1, it consists of several convolutional layers and blocks designed to capture high-level spatial features while preserving image details. The process starts with an initial convolutional layer for feature extraction, followed by multiple convolutional blocks. Each block contains two layers: the first uses a stride of 1 to maintain spatial resolution, and the second uses a stride of 2 to downsample the feature map. This reduces dimensionality while capturing higher-level features.

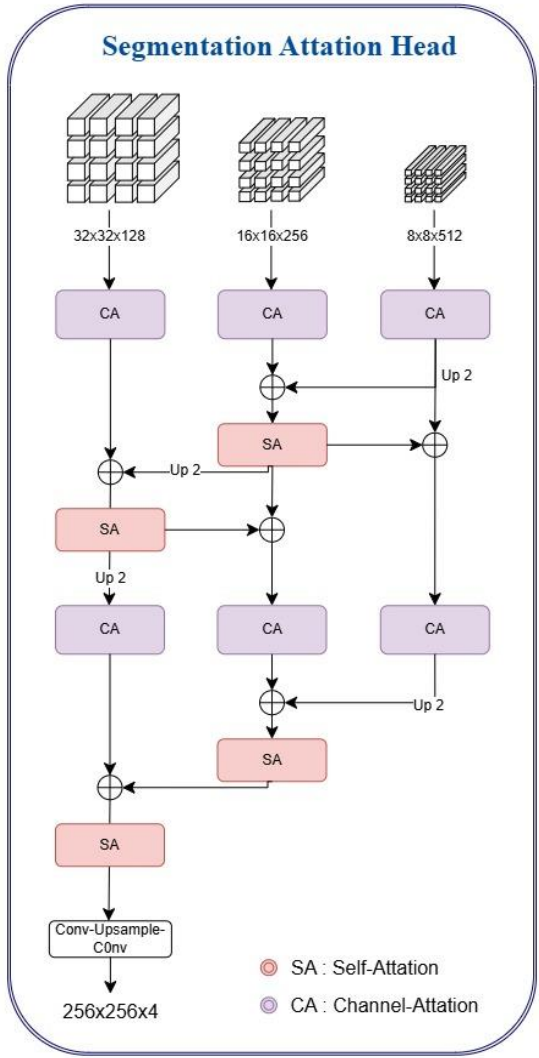

Fig. 2. Segmentation Attention-based Head (SAH)

To improve generalization, Group Normalization (GN) [ref] is applied after each block, stabilizing the training process, especially for medical images with varying characteristics. ReLU (Rectified Linear Unit) activation is used after each convolution to introduce non-linearity, enabling the network to capture complex patterns. The Backbone consists of five sequential convolutional blocks, working together to extract hierarchical features for the next stages of the network.

## B. YOLOv11 Neck

In this study, we leveraged the YOLO11[18] neck to design our model, taking full advantage of its advanced feature fusion and refinement capabilities to achieve state-of-the-art results in our specific application domain. The neck in YOLO11 is a pivotal component that bridges the backbone (feature extractor) and the head (task-specific output layers), playing a crucial role in refining and aggregating multi-scale feature maps for accurate object detection. It incorporates advanced modules like the C3K2 and C2PSA which enhance feature representation and computational efficiency, respectively. The Feature Pyramid Network (FPN) and Path Aggregation Network (PAN) work together to fuse features from different scales, ensuring the model can detect objects of varying sizes effectively. Additionally, the SPFF (Spatial Pyramid Pooling Fast) module improves multi-scale feature extraction, enabling better handling of complex scenes. These innovations make YOLO11's neck more efficient and accurate compared to previous versions, contributing significantly to its superior performance in real-time object detection, segmentation, and other computer vision tasks. By integrating the YOLO11 neck into our architecture, we were able to enhance our model's ability to process and interpret complex visual data, achieving remarkable accuracy and efficiency in our experiments.

## C. Segmentation Attention-based Head (SAH)

The (SAH), as shown in Fig. 2, is designed to enhance segmentation accuracy by integrating Self-Attention (SA) and Channel-Attention (CA) mechanisms, effectively refining features across multiple spatial scales. It processes multi-scale input features (32×32×128, 16×16×256, and 8×8×512) by first applying Channel Attention (CA) modules to emphasize important channels at each scale. Smaller-scale features are progressively upsampled and fused with higher-resolution features through addition, enabling seamless multi-scale integration. After each fusion, Self-Attention (SA) modules refine spatial relationships to further improve segmentation accuracy. At the final stage of the SAH, a combination of convolution, upsampling, and another convolution operation is applied to the refined features, ensuring the output is transformed into the desired resolution of 256×256×4. By leveraging both spatial and channel attention mechanisms and maintaining smooth transitions through convolution and upsampling, the SAH delivers precise and efficient segmentation performance tailored for complex visual scenes.

## 2. Loss Function

In each iteration, the backpropagation process is conducted using a newly proposed weighted multi-class exponential loss function (WME). This loss function is inspired by [28] designed for binary segmentation problems. Similar to earlier loss functions, the WME loss operates at the pixel level. It focuses on addressing the class imbalance problem in multi-class tasks by assigning different weights to each class, giving higher weights to minority classes compared to majority classes. To achieve this, we define a weighted term that adjusts the importance of each class. Let X represent an image with dimensions H×W×C, where H, W, and C are the height, width, and channels of the image, respectively. Y is the corresponding mask with dimensions H×W×K, where K is the number of classes. The goal is to train a segmentation model that maps X to Y, where Y contains class labels {0,1,..,K} for each pixel based on the labeled data.

The proposed WME loss is calculated as follows:

$$Loss = \sum_{h=1}^{H} \sum_{w=1}^{W} \left( \lambda \cdot \beta_1 \cdot exp^{-y_{true}^{h,w} * y_{pred,i}^{h,w}} + \beta_2 \cdot exp^{\left((1-y_{true}^{h,w}) \times \sum_{j \neq i}^{K} y_{pred,j}^{h,w}\right)} \right) \quad (1)$$

$$\lambda = exp^{-cr} \quad (2)$$

λ is a weighting parameter calculated based on the class imbalance ratio using E.q. 2.

where "cr" is the class rate, defined as the ratio of the number of pixels in a class to the total number of pixels across all classes. Lower class rates (cr) result in higher weights, making the loss function more sensitive to minority classes. $y_{true}$ is ground truth label for a pixel, where y = 1 for the true class and y = 0 for all other classes. (1−$y_{true}$) represents the complement of the ground truth and its value is 1. y $_{pred}$ is the probability that a pixel belongs to the true class i, and $\sum_{j \neq i}^{K} y_{pred,j}$ represents the sum of probabilities for all other classes. These probabilities are obtained from the segmentation network using the softmax function. β1 and β2 are scaling factors that control the importance of positive and negative contributions to the loss function. where we set β1 = 2 and β2 = 1 to appropriately balance these contributions. We show the convexity of our loss function in Appendix 1.

IV. EXPERIMENTS

In this section, we present the experimental setups, which encompass the datasets used and implementation details, including parameter configurations and evaluation metrics.

A. Data set

The Automated Cardiac Diagnosis Challenge (ACDC) dataset consists of data from 150 patients, acquired using cine-MR scanners. It is divided into 100 volumes with human annotations and 50 volumes reserved for private evaluation.

For our experiments, we use the 100 annotated volumes, splitting them into 60 samples for training and 30 samples for testing and 10 samples for validation..

B. Implementation Details

We conducted our experiments using the TensorFlow framework on an RTX 3070Ti GPU. The training used a batch size of 8, with images resized to 256×256. ReLU activation was applied after batch normalization for improved performance. For optimization, we employed the Adam optimizer and a "poly" learning rate policy for the ACDC datasets, where the learning rate decayed as $\left(1 - \frac{iter}{max_{iter}}\right)^{power}$, with an initial learning rate of 0.01 and power of 0.9. A weight decay of 0.0001 was used to prevent overfitting. To evaluate model performance, we use Dice and IoU metrics, common in segmentation tasks.

V. RESULT

This section presents both quantitative and qualitative results to evaluate the performance of our model. The quantitative results in Table 1 show the Dice Similarity Coefficients (DSC) for the segmentation of the right ventricle (RV), myocardium (Myo), and left ventricle (LV) across different methods. In addition, we provide qualitative results through visual examples of the model's segmentation outputs, illustrating its ability to accurately delineate cardiac structures. Furthermore, we demonstrate the convergence of our proposed weighted multi-class loss function (WME), highlighting its effectiveness in stabilizing the training process and improving performance over iterations.

TABLE I. COMPARISON OF DIFFERENT METHODS ON ACDC. THE BEST RESULTS ARE BOLDED.

| Methods | DSC↑ | | | |
|---|---|---|---|---|
| | AVG | RV | Myo | LV |
| R50 U-Net[10] | 87.55 | 87.10 | 80.63 | 94.92 |
| R50 Att-UNet[20] | 86.75 | 87.57 | 79.20 | 93.92 |
| R50 ViT[24] | 87.57 | 86.07 | 81.88 | 94.75 |
| TransUne[21] | 89.71 | 88.86 | 84.53 | 95.73 |
| SwinUnet[25] | 90.00 | 88.55 | 85.62 | 95.83 |
| LeViT-UNet-384[27] | 90.32 | 89.55 | 87.64 | 93.76 |
| YM-WML | 91.02 | 90.47 | 87.73 | 94.84 |

Table 1 shows the segmentation performance of various methods on the ACDC dataset, using the Dice Similarity Coefficient (DSC) for the right ventricle (RV), myocardium (Myo), and left ventricle (LV). The YM-WML method achieved the highest overall DSC of 91.02, with strong performance across all structures: 90.47 for RV, 87.73 for Myo, and 94.84 for LV. SwinUnet (90.00) and LeViT-UNet-384 (90.32) also performed well but slightly underperformed compared to YM-WML. R50 U-Net and R50 ViT had lower DSC scores of 87.55 and 87.57, respectively, showing better performance in LV but weaker results for RV and Myo. In conclusion, YM-WML outperformed other methods, achieving the highest DSC across all heart chambers.

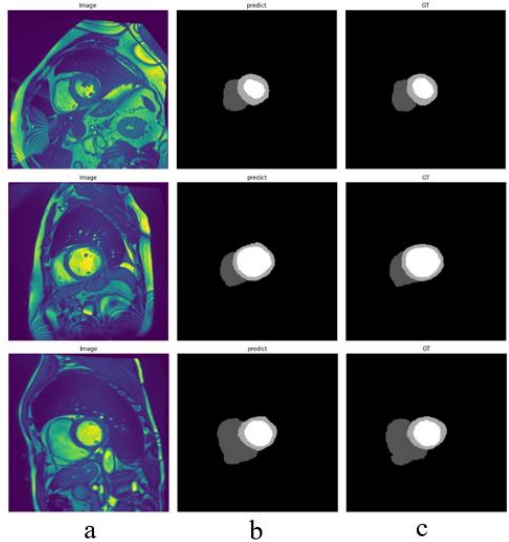

Fig. 3. Segmentation performance of the proposed YM-WML model on the ACDC dataset.

As illustrated in Figure 3, the comparison includes the original medical images (a), the predicted segmentations by our model (b), and the ground truth (c). The figure highlights the model's accuracy and consistency, with predictions closely aligning with the ground truth, demonstrating its effectiveness in segmenting cardiac structures across various cases.

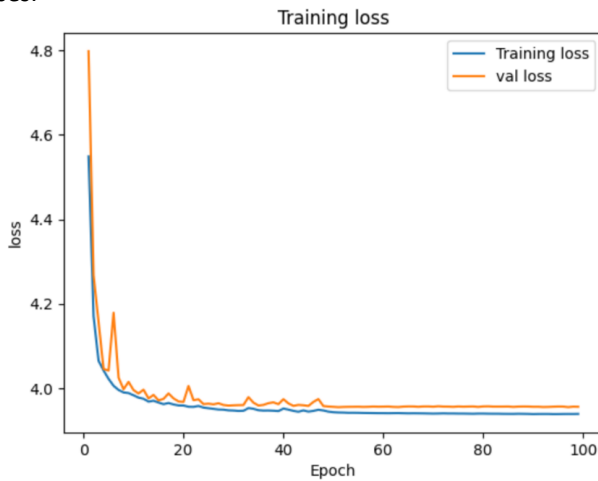

Fig. 4. The convergence rate of the loss function through ACDC dataset, varying with the epoch number

## VI. CONCLUSION

This study introduced the YM-WML model, which achieved a Dice Similarity Coefficient of 91.02 on the ACDC dataset, surpassing state-of-the-art methods. The model utilizes a robust backbone for feature extraction, a YOLOv11 neck for multi-scale feature aggregation, and an attention-based head for precise segmentation. The Weighted Multi-class Exponential (WME) loss further addressed class imbalance, ensuring robust and accurate cardiac segmentation, demonstrating YM-WML's potential for clinical applications.

REFERENCES

[1] M.A. Al-Masni, D.-H. Kim, CMM-Net: contextual multi-scale multi-level network for efficient biomedical image segmentation, Sci. Rep. 11 (2021) 1–18.

[2] K. Wang, X. Zhang, X. Zhang, Y. Lu, S. Huang, D. Yang, EANet: iterative edge attention network for medical image segmentation, Pattern Recogn. 127 (2022) 108636.

[3] D.R. Sarvamangala, R. V Kulkarni, Convolutional neural networks in medical image understanding: a survey, Evol. Intell. 15 (2022) 1–22.

[4] S. Ghosh, N. Das, I. Das, U. Maulik, Understanding deep learning techniques for image segmentation, ACM Comput. Surv. 52 (2019) 1–35.

[5] Zhou T, Ruan S, Canu S. A review: Deep learning for medical image segmentation using multi-modality fusion. Array. 2019 Sep 1;3:100004.

[6] Y. Abakarim, M. Lahby, A. Attioui, Bank failure prediction: a deep learning approach, Proc. 13th Int. Conf. Intell. Syst. Theor. Appl. (2020) 1–7.

[7] A.A.A. Alshawi, J. Tanha, M.A. Balafar, An Attention-Based Convolutional Recurrent Neural Networks for Scene Text Recognition, IEEE Access, 2024, https:// doi.org/10.1109/ACCESS.2024.3352748.

[8] M. Salehpanah, J. Tanha, Z. Jafari, S.E. Roshan, S. Rezaei, Cardiology disease diagnosis by analyzing histological microscopic images using deep learning, 2023 13th, Int. Conf. Comput. Knowl. Eng. ICCKE 2023 (2023) 354–361, https://doi. org/10.1109/ICCKE60553.2023.10326275.

[9] J. Long, E. Shelhamer, T. Darrell, Fully convolutional networks for semantic segmentation, in: Proc. IEEE Conf. Comput. Vis. Pattern Recognit., 2015, pp. 3431–3440.

[10] O. Ronneberger, P. Fischer, T. Brox, U-net: convolutional networks for biomedical image segmentation, in: Med. Image Comput. Comput. Interv, 2015 18th Int. Conf. Munich, Ger, 2015, pp. 234–241. Oct. 5-9, 2015, Proceedings, Part III 18.

[11] Li, X., Li, M., Yan, P., Li, G., Jiang, Y., Luo, H. and Yin, S., 2023. Deep learning attention mechanism in medical image analysis: Basics and beyonds. International Journal of Network Dynamics and Intelligence, pp.93-116.

[12] Redmon J. You only look once: Unified, real-time object detection. InProceedings of the IEEE conference on computer vision and pattern recognition 2016.

[13] M. M. Rahman and R. Marculescu, "Medical image segmentation via cascaded attention decoding," in Proceedings of the IEEE/CVF Winter Conference on Applications of Computer Vision, 2023, pp. 6222-6231.

[14] A. Sahafi, A. Koulaouzidis, and M. Lalinia, "Polypoid lesion segmentation using YOLO-V8 network in wireless video capsule endoscopy images," Diagnostics, vol. 14, no. 5, p. 474, 2024.

[15] S. Jadon, A survey of loss functions for semantic segmentation, in: 2020 IEEE Conf. Comput. Intell. Bioinforma. Comput. Biol., 2020, pp. 1–7.

[16] H. Kervadec, J. Bouchtiba, C. Desrosiers, E. Granger, J. Dolz, I. Ben Ayed, Boundary loss for highly unbalanced segmentation, in: Int. Conf. Med. Imaging with Deep Learn, 2019, pp. 285–296.

[17] M. Yeung, E. Sala, C.-B. Schonlieb, ¨ L. Rundo, Unified focal loss: generalising dice and cross entropy-based losses to handle class imbalanced medical image segmentation, Comput. Med. Imag. Graph. 95 (2022) 102026.

[18] Glenn Jocher and Jing Qiu. Ultralytics yolo11, 2024.

[19] Andrew G Howard, Menglong Zhu, Bo Chen, Dmitry Kalenichenko, Weijun Wang, Tobias Weyand, Marco Andreetto, and Hartwig Adam. Mobilenets: Efficient convolutional neural networks for mobile vision applications. arXiv preprint arXiv:1704.04861, 2017.

[20] Ozan Oktay, Jo Schlemper, Loic Le Folgoc, Matthew Lee, Mattias Heinrich, Kazunari Misawa, Kensaku Mori, Steven McDonagh, Nils Y Hammerla, Bernhard Kainz, et al. Attention U-Net: Learning where to look for the pancreas. arXiv preprint arXiv:1804.03999, 2018.

[21] Jieneng Chen, Yongyi Lu, Qihang Yu, Xiangde Luo, Ehsan Adeli, Yan Wang, Le Lu, Alan L Yuille, and Yuyin Zhou. TransUNet:

Transformers make strong encoders for medical image segmentation. arXiv preprint arXiv:2102.04306, 2021.

[22] Ashish Vaswani, Noam Shazeer, Niki Parmar, Jakob Uszkoreit, Llion Jones, Aidan N Gomez, Lukasz Kaiser, and Illia Polosukhin. Attention is all you need. In Advances in neural information processing systems, pages 5998–6008, 2017.

[23] Fabian Isensee, Paul F Jaeger, Simon AA Kohl, Jens Petersen, and Klaus H Maier-Hein. nnu-net: a selfconfiguring method for deep learning-based biomedical image segmentation. Nature methods, 18(2):203–211, 2021.

[24] Alexey Dosovitskiy, Lucas Beyer, Alexander Kolesnikov, Dirk Weissenborn, Xiaohua Zhai, Thomas Unterthiner, Mostafa Dehghani, Matthias Minderer, Georg Heigold, Sylvain Gelly, et al. An image is worth 16x16 words: Transformers for image recognition at scale. arXiv preprint arXiv:2010.11929, 2020.

[25] Ze Liu, Yutong Lin, Yue Cao, Han Hu, Yixuan Wei, Zheng Zhang, Stephen Lin, and Baining Guo. Swin transformer: Hierarchical vision transformer using shifted windows. In Int. Conf. Comput. Vis., pages 10012–10022, 2021.

[26] Zhengzhong Tu, Hossein Talebi, Han Zhang, Feng Yang, Peyman Milanfar, Alan Bovik, and Yinxiao Li. Maxvit: Multi-axis vision transformer. In Eur. Conf. Comput. Vis., pages 459–479. Springer, 2022.

[27] Xu G, Zhang X, He X, Wu X. Levit-unet: Make faster encoders with transformer for medical image segmentation. InChinese Conference on Pattern Recognition and Computer Vision (PRCV) 2023 Oct 13 (pp. 42-53). Singapore: Springer Nature Singapore.

[28] Roshan S, Tanha J, Zarrin M, Babaei AF, Nikkhah H, Jafari Z. A deep ensemble medical image segmentation with novel sampling method and loss function. Computers in Biology and Medicine. 2024 Apr 1;172:1083.

## Appendix 1.

**Theorem:** Suppose function $f: \mathbb{R}^n \to \mathbb{R}$ is twice differentiable over an open domain. Then f is convex if its second derivative satisfies $f''(x) >= 0$ for all x in its domain.

**Proof:** We prove that the first and second term of our loss function are convex.

**-Step1: First term convexity**

We compute the first derivative of first term of Eq.(1) ($T_1$) with respect to θ. For simplicity we denote $y_{true}^{h,w} * y_{pred,i}^{h,w}$ as $y_{true}\, y_{pred}$.

$$\frac{dT_1}{d\theta} = -y_{true} \frac{dy_{pred}}{d\theta} e^{-y_{true} y_{pred}}$$

Taking the second derivative:

$$\frac{d^2 T_1}{d\theta^2} = y_{true} \frac{d^2 y_{pred}}{d\theta^2} e^{-y_{true} y_{pred}} + y_{true}^2 \left(\frac{dy_{pred}}{d\theta}\right)^2 e^{-y_{true} y_{pred}}$$

For, convexity, we require $\frac{d^2 T_1}{d\theta^2} \geq 0$. This holds if:

$$y_{true} \frac{d^2 y_{pred}}{d\theta^2} e^{-y_{true} y_{pred}} + y_{true}^2 \left(\frac{dy_{pred}}{d\theta}\right)^2 e^{-y_{true} y_{pred}} \geq 0$$

Since $e^{-y_{true} y_{pred}} > 0$ and $y_{pred}$ is a softmax function (which is always positive) the second derivative is always positive, ensuring convexity.

**-Step2: Second term convexity**

Similar to step 1, the first derivative of the second term ($T_2$) is:

$$\frac{dT_2}{d\theta^2} = (1 - y_{true}) \frac{dy_{pred}}{d\theta} e^{(1-y_{true}) y_{pred}}$$

Taking the second derivative, we will have:

$$\frac{d^2 T_2}{d\theta^2} = (1 - y_{true}) \frac{d^2 y_{pred}}{d\theta^2} e^{(1-y_{true}) y_{pred}} + (1 - y_{true})^2 \left(\frac{dy_{pred}}{d\theta}\right)^2 e^{(1-y_{true}) y_{pred}}$$

As in the first term, it is obvious that the second derivative of second term is also positive.

Finally, since both second derivatives are positive, their sum is also positive, proving that the total loss function is convex.